# Tracking Single-Cells in Overcrowded Bacterial Colonies*

Athanasios D. Balomenos, *Student Member, IEEE*, Panagiotis Tsakanikas, *Member, IEEE*, and
Elias S. Manolakos, *Senior Member, IEEE*

*Abstract*—Cell tracking enables data extraction from time-lapse "cell movies" and promotes modeling biological processes at the single-cell level. We introduce a new fully automated computational strategy to track accurately cells across frames in time-lapse movies. Our method is based on a dynamic neighborhoods formation and matching approach, inspired by motion estimation algorithms for video compression. Moreover, it exploits "divide and conquer" opportunities to solve effectively the challenging cells tracking problem in overcrowded bacterial colonies. Using cell movies generated by different labs we demonstrate that the accuracy of the proposed method remains very high (exceeds 97%) even when analyzing large overcrowded microbial colonies.

## I. INTRODUCTION

Data analysis of time lapse microscopy "cell movies" is an important tool allowing us to "zoom in" and observe dynamic biological processes at the single-cell level [1]. Recent studies have noted its importance for investigating how stochasticity (biological "noise") affects gene regulation, aspects of cell growth, cell proliferation etc. [2]. Mathematical models are important to form and test hypotheses for such phenomena [3]. Time-lapse movies can provide an abundance of time course data, extremely valuable for mathematical models' calibration and validation. However, the accurate, automated segmentation and tracking of individual cells, as they grow, move and divide in expanding bacterial colonies, remain major challenges [4]. Manual cell counting and tracking across frames is extremely laborious and error prone. Therefore, automation strategies are essential before we can add time-lapse image analysis in the arsenal of high throughput methods for systems microbiology.

Finding correspondences, or matches, between objects across successive image frames is a fundamental problem in computer vision [5] and video compression [6]. The problem becomes more complicated if an unknown transformation deforms the objects in different frames. This is often the case in time-lapse cell movies, since cells grow, proliferate, and push each other! Establishing cells correspondence across frames can become very complicated, especially when the frame rate is low and cells move a lot across frames. In addition, limitations of image capturing and pre-processing (segmentation) introduce deformations in extracted cell curvatures, making cell matching even harder.

Several software packages support the segmentation and tracking of cells in time lapse movies. Among them we mention TLM-Tracker [7], CellTracer [8], and Schnitzcells [9]. TLM-Tracker [7] employs two overlap based algorithms for tracking, namely overlapping boxes and overlapping regions, and allows users to choose among them based on the tracking problem complexity. CellTracer [8] uses neighboring cells information to compute likelihood scores for cells' identity between successive time steps and then applies an integer programming based method to generate cell correspondences and construct the colony's lineage tree. Schnitzcells [9] segments cells and tracks them in a frame-to-frame manner using an energy function optimization method [10]. However, all the aforementioned tools suffer from several limitations, the most important being, (i) lack of tracking automation and, (ii) lack of accuracy in overcrowded regions. They often require intense human involvement to be able to track cells in frames with considerable cell movement and/or cells overpopulation.

We introduce a new, fully automated approach overcoming the above limitations. In conjunction with our accurate cell segmentation algorithms that proceed cell tracking (to be presented elsewhere), it enables high throughput analysis and efficient estimation of single-cell properties in growing microbial communities, thus forming the basis for the development of a single-cell micro-environment analytics platform. Besides their robustness, even in overcrowded micro-colonies, the proposed methodology offers several new capabilities: tracking of multiple micro-colonies in the field of view, lineage trees construction for each micro-colony, visualization on the tree of single-cell properties as they evolve in time (e.g. cell length, area, distance from the colony's centroid, GFP intensity, etc.), visualization of cell tracks across frames etc. To the best of our knowledge, our approach, which is inspired by motion estimation for video compression [6] [11], is the only one based on a dynamic ad hoc cell neighborhood formation and optimal matching following a divide and conquer strategy.

The rest of the paper is organized as follows. In Section II we present an overview of the developed tracking strategy. In Section III we show that it is accurate and outperforms

*This work was supported by the action THALIS-BIOFILMS, co-financed by EU (European Social Fund-ESF) and Greek national funds through the Operational Program "Education and Lifelong Learning" of the National Strategic Reference Framework (NSRF)-Research Funding Program: THALES. Investing in knowledge society through the ESF.

A. D. Balomenos is with the Informatics and Telecommunications Dept., University of Athens, Greece (e-mail: abalomenos@di.uoa.gr).

P. Tsakanikas is with the Food Science and Technology Dept., Agricultural University of Athens, Greece (e-mail: tsakanik@di.uoa.gr).

Prof. E. S. Manolakos is with the University of Athens, Greece (e-mail: eliasm@di.uoa.gr) and Visiting Scholar, Wyss Institute of Biologically Inspired Engineering, Harvard University.

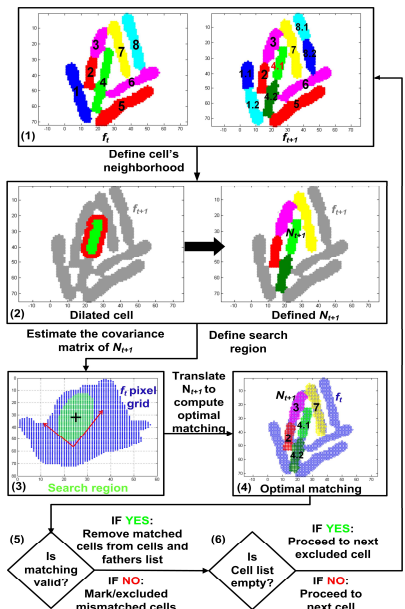

Figure 1. Schematic overview of the proposed tracking methodology. (1) Cells correspondence in frame $t$ ($f_t$) and frame $t+1$ ($f_{t+1}$), (2) Dilation of cell $c$ and cell neighborhood definition, (3) Definition of the search region in $f_t$, (4) Optimal matching, (5) Matching validity check.

state-of-the art methods when colonies become overcrowded. Finally, in Section IV we summarize our findings and point to work in progress.

## II. CELL TRACKING APPROACH

### A. The general idea

After cells segmentation is completed and in order to construct the lineage tree of a colony, we need to solve the cell tracking problem. Suppose that a colony in frame $t$ (to be called $f_t$ from now on) has $m$ cells and in the next frame $t+1$ ($f_{t+1}$) it has $n$ cells. Cell correspondence relations (matches) can be: 1-to-1 (father-to-daughter; proliferation), 1-to-2 (father-to-daughters; division), and even 1-to-$N$, $N>2$ (father to $N$ cells/objects; e.g. over-segmentation error). The general idea of our approach is first for every unmatched cell $c$ in the current frame $f_{t+1}$ to define an appropriate neighborhood of cells. Then using the covariance structure of this neighborhood to define a search region in the previous frame $f_t$, and search inside it for the best matching of the cell neighborhood's image, in order to establish an optimal correspondence of cells in the neighborhood. Finally the algorithm validates the cells matching result before repeating the same procedure until all cells are matched (have been assigned a father). Our method is inspired by motion estimation [11], a basic operation for video compression [6].

### B. Processing Stages

The objective of the algorithm is to locate candidate fathers (cells in $f_t$) for each daughter cell $c$ of the current frame $f_{t+1}$. In the sequel we focus w.l.o.g. on a cell $c$ in $f_{t+1}$ to describe how the algorithm works. For example in Fig. 1 Panel 1 right cell $c=4.1$ (in $f_{t+1}$) should be matched to its father cell 4 (in $f_t$).

**Stage 1. Cell neighborhood Definition**: For each cell $c$ in the current frame $f_{t+1}$ we find unmatched cells with centroids inside a hypothetical disk centered at $c$ with radius $R$ equal to the average length of cells in $f_{t+1}$. For example let's assume that for cell $c=4.1$ this disk includes all colored cells (in $f_{t+1}$), see Fig. 1 Panel 1 right. The algorithm then dilates cell $c$ by using a disk structuring element [12] of radius $r$ equal to the average cell width (see Fig. 1 Panel 2 left). Then it identifies the cells "touched" by the dilated cell $c$. These cells are the first order neighbors of $c$ (Fig. 1 Panel 2 right). Then, it may apply the same dilation procedure again to the first order neighbors in order to find the second order neighbors of $c$, and so on. This recursion is repeated $L$ times, where $L$ is an upper bound for the layers of neighbors considered, resulting in the definition of the neighborhood of $c$ (called $N_{t+1}$). The value of $L$ depends on the location of $c$ and the size of the colony, and is larger for cells close to the colony's centroid.

**Stage 2. Search area definition** (Fig. 1 Panel 3): In order to match efficiently the defined neighborhood $N_{t+1}$ with candidate neighborhoods within the previous frame $f_t$, we should first define an appropriate search area $S_t$ in $f_t$. Initially we estimate the covariance of the pixels matrix of cells in $N_{t+1}$ and use it to compute the Mahalanobis distance [13] of each pixel of $f_t$ to the image of the centroid of $N_{t+1}$ into $f_t$ (see the cross in Fig. 1, Panel 3 left). Then we find the $k$ nearest neighbor (kNN) pixels [13] to the centroid image (green area Fig. 1, Panel 3 left). Finally we select uniformly a subset of $l$ points, among the kNNs, to form the set of points, $S_t$, for the candidate centroids of the best neighborhood in $f_t$ i.e. the one matching optimally $N_{t+1}$. We remark that parameters $l$ and $k$ take values proportional to the size of the size of $N_{t+1}$ in pixels. As it is apparent the algorithm exploits the orientation of $N_{t+1}$ (covariance structure) to estimate the direction of its motion between consecutive frames and thus constrain the number of candidate matchings to be evaluated.

**Stage 3. Compute/Evaluate the candidate matchings:** We now place the image of the centroid of $N_{t+1}$ to each point in $S_t$ and create $l$ score matrices, $S_l$, having $|N_{t+1}|$ rows and $m$ columns each. The $(i,j)$-th element of a score matrix is the *overlap score* of the i$^{th}$ cell in $N_{t+1}$ and the j$^{th}$ cell of $f_t$ defined as:

$$S_{ij} = \text{area}(c_{(i,t+1)} \cap c_{(j,t)})/\text{area}(c_{(i,t+1)} \cup c_{(j,t)}). \quad (1)$$

Then for each $S_l$ we also compute its *overall overlap score*

$$O_l = \sum_{1 \le i \le |N_{t+1}|} \max_j (S_{ij}), \quad (2)$$

i.e. the sum of maximum overlap scores of each cell in $N_{t+1}$ (i$^{th}$ row) with a cell in $f_t$ (j$^{th}$ column). This maximum establishes a candidate correspondence between each cell $i$ in $N_{t+1}$ (daughter cell) to one and only one cell $j$ in $f_t$ (father cell). When considering all rows of the score matrix this leads to a candidate matching of all cells of neighborhood $N_{t+1}$ to father cells in the previous frame $f_t$.

**Stage 4. Determine the optimal matching** (Fig.1 Panels 4): We choose the candidate neighborhood in $f_t$ with the highest overall score $O_l$, to be called the $N_t$. Then we create a new matrix $\hat{S}$ which has $|N_{t+1}|$ rows and $|N_t|$ columns and its elements are defined as:

$$\hat{S}_{ij} = \begin{cases} S_{ij} & \text{if cell } i \text{ corresponds to cell } j \\ 0 & \text{otherwise} \end{cases} . \quad (3)$$

The nominal case for this matrix is to contain columns with one or two non-zero elements, because each father cell should correspond to at most two daughter cells.

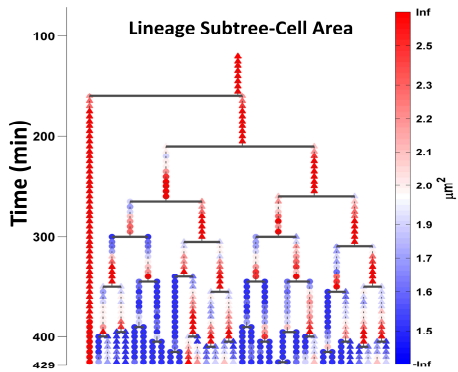

**Figure 2.** Lineage Tree Visualization of cell area evolving with time as cells grow and divide. Triangular (circular) nodes depict time points (resolution 5 min) in the life of an external i.e. on micro-colony's boundary (internal) cell.

**Stage 5. Validate the optimal matching** (Fig.1 Panels 5-6)**:** Considering the aforementioned expected nominal behavior, we assess the validity of the optimal matching by estimating the *total overlap score* for each father cell $j$ in $N_t$:

$$O_j = \sum_{1 \leq i \leq |N_{t+1}|} \hat{S}_{ij} . \qquad (4)$$

If for every cell $j=1,2,...|N_t|$, the score $O_j$ is greater than a threshold $T$ and less than 1 (maximum overlap) the optimal matching is accepted; *all* cells in $N_{t+1}$ and $N_t$ are considered matched and removed from the cells-to-match and fathers' lists respectively. Otherwise, the optimal matching is rejected and all cells in $N_{t+1}$ are marked as "problematic". If the same cell has been marked repeatedly (e.g. 3 times) it is removed from the cells-to-match list and placed in an exclusion list. This scheme allows us to continue the processing while also separating difficult cells-to-match cases, usually cells that were divided or moved radically between consecutive frames. The algorithm will revisit these "problematic" cells and try to find their fathers again at the end of the process, when the problem has become simpler, i.e. when the large majority of cells have been matched.

We repeat this process until each cell in $f_{t+1}$ is either matched with a cell in $f_t$ or removed from the cells-to-match list. At this point we try again to match the excluded "problematic" cells by following the same five-stage process. If this fails to match all remaining cells, we lower the threshold $T$, reset the excluded cells list and repeat the same process. Finally, when all cells in two consecutive frames are matched, or $T$ is down to 0.5, the algorithm proceeds with the next pair of frames until all pairs have been processed. Any unmatched cells at this point are most probably over-segmentation artifacts.

### C. Construction of the Lineage and Division Trees

As the algorithm tracks cells across frames it simultaneously creates a lineage tree, keeping record of the attributes of each individual cell (see Fig. 2). When a tracking step is completed, the algorithm searches the tree to find the father of each matched cell and inserts a new node under it. At the end, the algorithm returns as many lineage trees as the number of cells in the initial frame. Given these lineage trees, our method generates recursively another useful tree structure, the so called *divisions' tree*. Division trees record only cell division events and each node depicts an individual cell's "life attribute" (e.g. the average cell length).

### III. RESULTS AND DISCUSSION

#### A. Evaluation

Two cell movies created by different labs were used in the comparative evaluation. The first movie starts with four *S. Typhimurium* cells which grow to become four discrete micro-colonies with ~200 cells each [3]. The second movie shows an *E. Coli* micro-colony of ~50 cells [9]. The tracking ground truth for both movies was determined by experts. In order to evaluate the proposed method, we compared its performance to that of Schnitzcells [9]. since this is the most recent software package and gives satisfactory tracking results for both movies. However, Schnitzcells failed to segment the first movie, so to evaluate the tracking methods fairly we provided as input to Schnitzcells the manually refined results of our segmentation algorithms (not discussed in this paper).

Evaluation was performed following two methods. First, we evaluated the two algorithms using a frame-based approach, similar to the one proposed in [14], based on the estimated *Tracker Detection Rate* (TDR) defined as:

$$\text{TDR}=\text{TP}/\text{GT}, \qquad (5)$$

where True Positives (TP) is the number of frames with no tracking errors (i.e. cell-to-cell correspondences that were undetected or non-existing) and ground truth (GT) is the total number of frames in the movie. As we can see in Figure 3(a), the proposed method exhibits very high TDR for both datasets, over 98.7%. Moreover, it outperforms Schnitzcells even when using their own movie. As we observe in Fig. 3(b), Schnitzcells made errors mainly in the last frames where the micro-colonies become overcrowded and tracking becomes very difficult. So, to investigate the overpopulation effects we focused in the last few frames (79-86) and evaluated the two algorithms using also a tracks based approach, a more strict variation of the one presented in [14][15]. Here we consider as ground truth (GT) tracks with trajectory and lifespan extending to, or beginning after, the 79[th] frame. Specifically, we estimated the *Error Rate* (ER) that is defined as:

$$\text{ER}=(\text{FAT}+\text{TDF})/\text{GT} , \qquad (6)$$

where *False Alarm Track* (FAT) is the number of non-existing but detected tracks (a track is considered non-existing when it differs at least in one time point from the ground truth), and *Track Detection Failure* (TDF) is the number of existing but undetected tracks. Again, in both movies the proposed method exhibited an extremely low ER, under 1%, and an advantage higher than 3.3% relatively to the best currently available cell tracking approach. Moreover, our method exhibits a very low ER even in highly overcrowded micro-colonies. It seems that the divide and conquer strategy we follow is better than global optimization methods in such situations. As cell numbers increase exponentially, it is more probable for global optimization methods to get trapped to local minima.

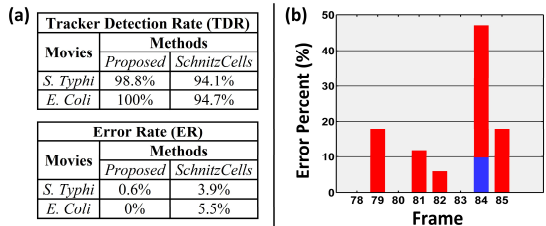
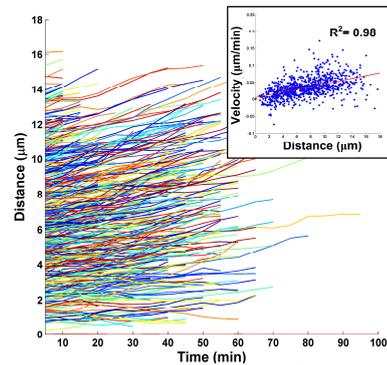

**Figure 3.** Evaluation. (a) *Top*: Frame based TDR of the methods under evaluation in *E. coli* and *S. Typhimurium* cell movies. *Bottom*: Error Rate for the last frames 79-85 with severe overcrowding. (b) Error distribution of Schnitzcells (red) and proposed method (blue) in the *S. Typhi.* movie. More than 50% of the errors occur in the last two frame pair matchings.

TABLE I. EVALUATION ON OVERCROWDED MICRO-COLONY

|  | TP | FAT | TDF | ACC | ER |
|---|---|---|---|---|---|
| *Proposed* | 911 | 21 | 21 | 97.7% | 4.5% |

Moreover, in order to check if the proposed method remains robust when the overcrowded colonies become very large we assessed its performance (Table I) considering two consecutive frames of a salmonella cell movie (frames 75 and 76, GT=932 cell matches). We observe that our method achieves very high Accuracy (over 97%) and very low ER (under 5%). Due to lack of space we provide in [16] the two successive image frames used (5 min. apart). We mark cells on the same track using the same color. Gray cells are those that were not matched correctly. We remark that most errors occur close to the large colony's boundaries, which indicates that our approach is indeed robust to severe over-population occurring at the colony's central region. Cells on the boundary exhibit usually higher mobility, so it is expected for some tracking errors to occur in the periphery, especially if the frame rate is small.

### B. Tracks Visualization

Figure 4 illustrates how our method can constitute a useful visual analytics tool to microbiologists. Here, we visualize with pseudo-color single-cell tracks allowing us to assess how each cell's distance from the colony's centroid varies with time during its lifespan. We observe that more distant cells exhibit, on average, higher mobility (higher track "slopes") than cells near the centroid. This is expected and conforms to physical rules since boundary cells can move unconstrained compared to internal cells. The scatter plot quantifies the correlation of cell's velocity to cell's maximum distance from the colony's centroid.

### IV. CONCLUSIONS AND FUTURE WORK

We presented a new divide-and-conquer cell tracking strategy inspired by block matching motion estimation for video compression. It can be used to track bacteria automatically and quantify at the single-cell level how their morphological and expression characteristics evolve with time. The algorithm is shown to outperform state of the art methods in overcrowded colonies. Moreover the single-cell attributes extracted from analyzed time-lapse movies can be visualized over lineage trees or cell track trajectories which can help microbiologists formulate new hypothesis for further experimental or modeling work.

**Figure 4.** Each colored line corresponds to a cell track depicting the cell's distance from the colony's centroid as a function of time (time-series). Max. distance from colony centroid and cell velocity are correlated (scatterplot, Spearman Correlation ≈ 0.54).

We currently work on combining cell segmentation and cell tracking algorithms into a closed loop system in order to improve their accuracy and robustness. The structure of the lineage tree can help us identify and correct segmentation errors (e.g. due to over-segmentation) which in turn can improve cell tracking. This is especially useful since there is no automatic way to assess cell segmentation quality, while there are several ways to assess tracking quality.

### REFERENCES

[1] J. C. W. Locke and M. B. Elowitz, "Using movies to analyse gene circuit dynamics in single cells," *Nat. Rev. Microbiol.*, vol. 7, no. 5, pp. 383-392, May 2009.

[2] N. Rosenfeld, TJ Perkins, U. Alon, MB Elowitz, PS Swain. A fluctuation method to quantify In Vivo Fluorescence Data. *Biophysical Journal*, 91:759-766, 2006.

[3] K. P. Koutsoumanis and A. Lianou, "Stochasticity in colonial growth dynamics of individual bacterial cells," *Appl. Environ. Microbiol.*, vol. 79(7):2294-2301, 2013.

[4] A. Elfwing, Y. LeMarc, J. Baranyi, and A. Ballagi, "Observing growth and division of large numbers of individual bacteria by Image Analysis," *Appl. Environ. Microbiol*, vol. 70, no. 2, pp. 675-678, Feb. 2004.

[5] A. Yilmaz, O. Javed, and M. Shah, "Object tracking: A Survey," *ACM Comput. Surv.*, vol. 38, no. 4, Jul. 2006.

[6] Xie Liyin; Su Xiuqin; Zhang Shun, "A review of motion estimation algorithms for video compression," Computer Application and System Modeling (ICCASM), 2010 International Conference on , vol.2, no., pp.V2-446,V2-450, 22-24 Oct. 2010

[7] J. Klein, S. Leupold, I. Biegler, R. Biedendieck, R. Münch, and D. Jahn, "TLM-Tracker: software for cell segmentation, tracking and lineage analysis in time-lapse microscopy movies," *Bioinformatics*, vol. 28, no. 17, pp. 2276-2277, Sep. 2012.

[8] Q. Wang, J. Niemi, C. M. Tan, L. You, and M. West, "Image segmentation and dynamic lineage analysis in single-cell fluorescence microscopy," *Cytometry A.*, vol. 77, no. 1, pp. 101-110, Jan. 2010.

[9] J. W. Young, J. C. W. Locke, A. Altinok, N. Rosenfeld, T. Bacarian, P. S. Swain, E. Mjolsness, and M. B. Elowitz, "Measuring single-cell gene expression dynamics in bacteria using fluorescence time-lapse microscopy," *Nature Protocols*, vol. 7, no. 1, pp. 80-88, Dec. 2011.

[10] V. Gor, T. Bacarian, M. Elowitz, and Eric Mjolsness, "Tracking Cell Signals in Fluorescent Images," in Workshop on Computer Vision in Bioinformatics, IEEE CVPR Annual Meeting, San Diego, Jul. 2005.

[11] A. Barjatya, "Block Matching Algorithms For Motion Estimation," *IEEE Trans. Evol. Comput.*, vol. 8, pp. 225-229, 2004.

[12] R. Gonzales, R. Woods, *Digital Image Processing,* Addison-Wesley, 2007.

[13] S. Theodoridis, K. Koutroumbas, "*Pattern Recognition*," 4th ed., Elsevier, 2008.

[14] F. Bashir and F. Porikli, "Performance evaluation of object detection and tracking systems," *IEEE Int'l Workshop on Performance Evaluation of Tracking and Surveillance (PETS 2006)*, New York, pp. 7-14, Jun. 2006.

[15] F. Yin, D. Makris, and S. A. Velastin, "Performance evaluation of object tracking algorithms," in Proc. IEEE Int. Workshop Perform. Eval. Tracking Surveillance, 2007, pp. 733–736.

[16] Tracking Single-Cells in Overcrowded Colonies: https://db.tt/f4MPbEbd